\pdfoutput=1
\documentclass[11pt]{article}

\usepackage[final]{coling}

\usepackage{times}
\usepackage[inkscapelatex=false]{svg}
\usepackage{latexsym}
\usepackage{amssymb}
\usepackage{float}
\usepackage{placeins}

\usepackage[T1]{fontenc}
\usepackage[utf8]{inputenc}

\usepackage{microtype}
\usepackage{multirow}
\usepackage{booktabs}
\usepackage{soul}

\usepackage{color-edits}
\addauthor[Alexis]{ap}{cyan}
\addauthor[Ali]{am}{teal}
\addauthor[Katharina]{kw}{blue}
\addauthor[Enora]{er}{violet}
\addauthor[Luke]{lg}{blue}

\usepackage{inconsolata}
\usepackage{graphicx}
\usepackage{subcaption} % Required for subfigures

\title{From Priest to Doctor:\\Domain Adaptation for Low-Resource Neural Machine Translation}

\author{
 \textbf{Ali Marashian,\textsuperscript{1}}
 \textbf{Enora Rice,\textsuperscript{1}}
 \textbf{Luke Gessler,\textsuperscript{2}}
 \\
 \textbf{Alexis Palmer,\textsuperscript{1}}
 \textbf{Katharina von der Wense\textsuperscript{1,3}}
\\
 \textsuperscript{1}University of Colorado Boulder,
 \textsuperscript{2}Indiana University Bloomington,
 \\
 \textsuperscript{3}Johannes Gutenberg University Mainz
\\
\href{mailto:email@domain}{ali.marashian@colorado.edu}
 }

\begin{document}
\maketitle

\begin{abstract}
Many of the world's languages have insufficient data to train high-performing general neural machine translation (NMT) models, let alone domain-specific models, and often the only available parallel data are small amounts of religious texts. Hence, domain adaptation (DA) is a crucial issue faced by contemporary NMT and has, so far, been underexplored for low-resource languages.
In this paper, we evaluate a set of methods from both low-resource NMT and DA in a realistic setting, in which we aim to translate between a high-resource and a low-resource language with access to only: a) parallel Bible data, b) a bilingual dictionary, and c) a monolingual target-domain corpus in the high-resource language.
Our results show that the effectiveness of the tested methods varies, with the simplest one, DALI, being most effective. We follow up with a small human evaluation of DALI, 
which shows that there is still a need for more careful investigation of how to accomplish DA for low-resource NMT.

% Our results - the first for application of these methods to this setting - show that the approaches fail to produce usable target-domain NMT systems, calling for more careful investigation of how to accomplish domain adaptation in the low-resource NMT setting.
\end{abstract}

\section{Introduction}

Neural machine translation (NMT) models have limited ability to deal with %low-resource 
languages that lack large-scale monolingual and parallel corpora %for model training
\cite{wang2021survey}. Moreover, NMT systems face challenges when translating text from novel domains characterized by unique style or vocabulary \cite{koehn-knowles-2017-six, saunders2022domain}. Often, these issues co-occur, a scenario that has been neglected by researchers so far. 
Most of the world's 7000+ languages are considered low-resource \cite{joshi2020state}, and existing data for them are in limited domains; the languages that could most benefit from domain adaptation (DA) are the ones left behind.

% \begin{figure}[t!]
%     \centerline{\includesvg[width=1\columnwidth]{drawing.svg}}
%     \caption{The neglected intersection of Low-Resource NMT and DA in NMT has led us to put these constraints in place: out-of-domain parallel corpus (e.g. the available religious texts), a bilingual lexicon, and in-domain, monolingual texts in the high-resource language.}
%     \label{fig: example}
% \end{figure}

% \begin{figure}[htbp]
%   \centering
%   \includesvg{tables/drawing.svg}
%   \caption{svg image}
% \end{figure}

\begin{figure}[t!]
    \centering
    \centerline{\includesvg[width=1\columnwidth]{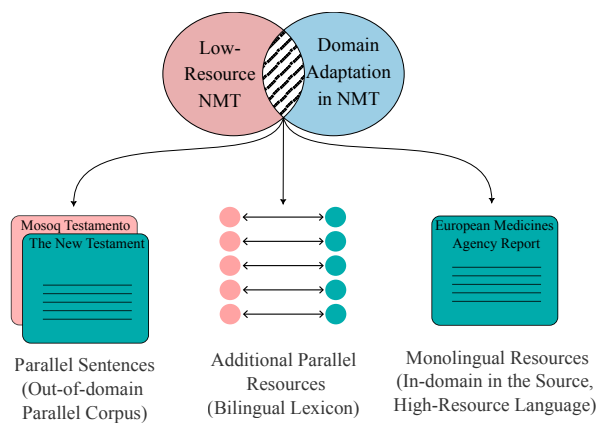}}
    %\caption{The (neglected) intersection of Low-Resource NMT and DA in NMT, has led us to consider only these commonly accessible resources.}
    \caption{In our work, which looks at the (previously neglected) intersection of low-Resource NMT and domain adaptation in NMT, we consider only these commonly accessible resources.}
    \label{fig: resources}
\end{figure}

In this paper, we explore a realistic setting in which we aim to translate between a high-resource and a low-resource language and are restricted to 
%utilize 
the following commonly available resources: a) Bible translations, i.e., a small parallel corpus in the source domain; b) monolingual target-domain texts in the high-resource language; 
%that is also in the domain of interest (target domain), 
and c) a bilingual dictionary for the two languages. 
%As we are interested in a 
To keep the setting generalizable, we assume neither access to a model pretrained on text in the low-resource language nor access to data in a related high-resource language, as for many truly low-resource languages, those are impossible to find.
%\amcomment{Many languages do not even have a related high-resource language to utilize here,\footnote{Like many languages of North and South America.\amcomment{TODO: need citation}} thus making the limited resources we have chosen a generalizable setting to many truly low-resource languages. As finding data in different domains for truly low-resource languages proved difficult, we \textit{simulate} these settings by using a base multilingual model that was not pre-trained on them. }
%that ideally covers the in-domain terms of the target domain. 

%\ermargincomment{With these limitations in place, we experiment with methods in translating from English to a simulated low-resource target language.}
% We probe this setting by inspecting some methods for domain adaptation when translating from English to some low-resource language.
We experiment with 
%different 
a set of four DA and low-resource NMT methods and aim to translate from English to a target language, simulating a low-resource setting. 
We use mBART \cite{liu-etal-2020-multilingual-denoising}
which has been fine-tuned on parallel Bible texts as our base model, and our goal is to adapt it to the target domains of government documents and medicine.
%Our (source-domain) parallel corpora are exclusively biblical verses, and our 
%For the 
%target domains 
%we examine both 
%are government and medical documents.
%, while the only parallel corpora we have access to are biblical verses. 
%We use mBART \cite{liu-etal-2020-multilingual-denoising} as the base model \lgmargincomment{combine with the sentence two before this to avoid redundancy} and implement some different methods on top of it. 
The methods we investigate use the bilingual dictionaries in various ways.
%All these methods use a bilingual dictionary, but all in different aspects. The target languages are not low-resource; however, they are not used in mBART's pre-training and can hence to some extent simulate an actual low-resource language. 
%\apmargincomment{Add description of experimental set-up/method/approach} \ammargincomment{Is this something?}

Our experiments showcase the varying effectiveness of existing methods: 
%for the realistic setting of building a domain-specific NMT system for a low-resource language without parallel domain-specific data: 
the weakest approach results in models that perform \emph{worse} than the base model, while the best approach -- which, surprisingly, is also the simplest -- results in a ChrF score more than twice as high as the base model's. However, as the best model only reaches a ChrF score of 42.47 and a BLEU score of 13.47 (on average), we also perform a small human evaluation, which confirms that there is still a need for the development of better DA methods for low-resource NMT. Our code is available on GitHub.\footnote{\href{https://github.com/alimrsn79/da_lr_nmt/}{https://github.com/alimrsn79/da\_lr\_nmt}}

%proposed setting. To the best of our knowledge, no existing work directly studies domain adaptation in NMT for low-resource languages in this setting. Thus, the methods either were proposed for "domain adaptation in NMT" or "low-resource NMT". 

\section{Related Work}

\paragraph{Domain Adaptation in NMT }
As domains are defined by the characteristics of data \cite{saunders2022domain}, many effective DA approaches focus on the data and, thus, 
%alone rather than the underlying  architecture \apmargincomment{be more specific}\ammargincomment{hope this is clearer}, as domains are defined by the characteristics of data \cite{saunders2022domain}. 
can be applied to various underlying architectures.
% One approach involves extending training data by retrieving general-domain examples that resemble the target-domain
% \cite{xu-etal-2020-boosting, poncelas-etal-2019-transductive, cai-etal-2021-neural, vu2021machine}. 
% A group of approaches 
Some works focus on acquiring monolingual in-domain data, which is easier
%is usually much more convenient 
to find than in-domain parallel data. Back-translation uses monolingual target-domain data in the
%in the target domain and 
target language and produces artificial source sentences using 
%relies on 
a target-to-source NMT model \cite{poncelas2019adaptation, jin2020simple}.
%When data in the target language and target domain is available, synthetic source-side data can be produced  There are some approaches that make 
\citet{chinea-rios-etal-2017-adapting} use monolingual source-side corpora and a source-to-target NMT model for forward-translation, 
where it is common to employ self-learning. 
% \cite{gordon-duh-2020-distill} use one or several stronger teacher models, training or tuning a smaller student model on their outputs. 
With access to a small parallel corpus, extra training data can be created by introducing noise \cite{vaibhav-etal-2019-improving}. Synthetic parallel data can be acquired from an external source or generated using a predefined or induced lexicon. \citet{dali} use a lexicon to back-translate target-side sentences. \citet{peng2020dictionary} use a dictionary,
%for data augmentation, 
injecting dictionary terms 
%at appropriate parts of 
into out-of-domain texts to synthesize in-domain training data. \citet{bergmanis-pinnis-2021-facilitating} augment the training data by annotating randomly selected source language words with their target language lemmas to integrate terms.
\citet{zhang-etal-2022-iterative} introduce lexical constraints into iterative back-translation.

Other approaches add parameters to the model, e.g., domain tags
%, introducing extra tag embeddings, or even submodules
\cite{kobus-etal-2017-domain,stergiadis-etal-2021-multi}. Such a manipulation of the embeddings could extend to more terms in the vocabulary, beyond the tags \cite{pham-etal-2019-generic,sato-etal-2020-vocabulary, man2023wdsrl}.
% Auxiliary trainable subnetworks could be added to the model \cite{lin-etal-2021-towards}.
%It could be in the form of 
With adapter-based methods, a domain-specific module is trained
%, freezing other parameters for more efficient training 
\cite{bapna-firat-2019-simple}. \citet{leca} use a pointer-generator to copy suggestions from the input, which come from a domain-specific dictionary. 

\paragraph{Low-Resource NMT }
Methods for low-resource MT show some overlap with DA methods.
 %, as a new language can be viewed as an extreme case of a new domain. 
 One popular approach is data augmentation, which can be in the form of word or phrase replacement with the help of a bilingual lexicon \cite{nag2020incorporating}. Back-translation, forward-translation, and data selection methods can also be applied \cite{sennrich-etal-2016-improving, fadaee-monz-2018-back, dou-etal-2020-dynamic}. Transfer learning is a useful technique in low-resource NMT \cite{10.1145/3314945, kocmi-bojar-2020-efficiently, cooper-stickland-etal-2021-recipes}. 
\citet{liu-etal-2021-continual} continue to pretrain mBART \cite{liu-etal-2020-multilingual-denoising} on unseen languages, %slightly changing the pretraining objective to 
utilizing a bilingual dictionary. 
% Some semi-supervised approaches include using monolingual data to create language models \cite{gulcehre2015using} and multi-task learning \cite{domhan-hieber-2017-using}. However, they require a large amount of monolingual texts which still be unfeasible for low-resource languages. 
Although we do not inspect large language models (LLMs) in our experiments, some recent works explore the potential of LLMs for low-resource NMT. \citet{robinson-etal-2023-chatgpt} observe that ChatGPT's MT capabilities across the 204 languages of the FLORES-200 dataset \cite{costa2022no} consistently lag behind traditional NMT models.
%in the case of low-resource languages. 
\citet{ghazvininejad2023dictionary} use dictionaries to suggest words %for the model 
to use in the output translation.
%give the model some suggestions for the words to use in the output translation, while giving the model some examples of the translation
\citet{zhang2024teaching} %improve their model by 
adopt different strategies for dictionary term lookup and the retrieval of examples for in-context learning.
%% ADD BACK FOR FINAL VERSION
\citet{siddhant2022towards, 10.1145/3567592} note that, in the case of many low-resource languages, the problem is more severe since the only available parallel data are religious texts.

\section{Data}
\label{sec:data}
\paragraph{Parallel Source-Domain Data } In all experiments, the only \emph{parallel} data we use for training come from the JHU Bible Corpus
%In all the experiments, we limit ourselves to The John Hopkins University Bible Corpus 
\cite{mccarthy-etal-2020-johns}. 
%as our only source of parallel data. 

\paragraph{Target-Domain Data} We explore adapting to two different domains, one at a time: government documents and medicine. The domain-specific data mostly come from past WMT translation tasks \cite{barrault-etal-2020-findings, akhbardeh-etal-2021-findings, kocmi-etal-2022-findings, kocmi-etal-2023-findings}. 
%for government and medical data. 
As we assume only \emph{monolingual} in-domain training data (cf. Section~\ref{sec:exp}), training and pretraining use only source-side sentences from these parallel data sets. %\footnote{We could of course use other monolingual data sources, but we use (mostly) WMT data for experimental flexibility.}
% \apmargincomment{The argument for not using external data wasn't really convincing - I made a different justification, but I'm not sure that this one is convincing either.}
Data availability varies across language/domain pairs, and we cap data set sizes to maintain comparability across languages.
For \textbf{training} we use no more than 200K sentence pairs.
If our setting requires \textbf{pretraining}, we use the same source-side sentences used for training.
For \textbf{testing}, we use 1500 sentence pairs.

%As will be discussed in Section \ref{sec:exp}, there are two stages we might need data for in different settings: training and pre-training, and in both cases we only need the source-side sentences. We use 1500 sentence pairs as the test split in all the settings, and no more than 200K sentence pairs for training (and if the setting does require pre-training, we use the same source-side sentences that we used for training for pre-training as well). %There are varying levels of availability of data for different domains in different languages, and that is why we do not use more than 200K sentences: for the different experiments to be roughly comparable. 
%Note that we can use some irrelevant source-side text in the target domain for both training and pre-training as well, but we hypothesized that using the source-side
%data from a corpus that translated already, might yield some more relevant data in regard to the target language and the people using that language; as the data was translated because it \textit{probably} needed to be translated. 
More details about the data used for each domain and language pair can be found in Appendix \ref{sec:app-data}.
%\lgmargincomment{this is a huge graf, try to break it up if you can}

\paragraph{Dictionaries }
\label{dict:making}
The methods we investigate here call for source--target language dictionaries.
To build dictionaries, for each language pair we extract the 5000 most frequent lemmas and their inflections from the monolingual training data and use the Google Translate API\footnote{\url{https://cloud.google.com/translate/}} to translate those words.\footnote{We expect performance might increase if we had domain-specific bilingual dictionaries for each language pair.}
%A bilingual dictionary is used for each of our approaches. We hypothesize that if the dictionary is in-domain we would see larger boosts in performance;
% This dictionary is preferably \lgmargincomment{you should clarify if this preference is just based on your intuitions or if the previous works you're reproducing say this} an in-domain one;%
%so if the target domain of translation is the medical domain, the dictionary would contain medical terms which would hopefully help the model with rare domain-specific vocabulary. 
%Instead of finding in-domain dictionaries for these target languages and domains, we extract the most frequent lemmas ($N=5000$) and their respective inflections on the source side of the training split, and use the Google Translate API\footnote{\href{https://cloud.google.com/translate/}{https://cloud.google.com/translate/}} to make up a dictionary.\lgmargincomment{If this is common practice for getting a dictionary like this it might be helpful for your presentation to say so}\ammargincomment{I don't think it is, but I did it cause it made sense. I don't know of any other paper that makes dictionaries with google translate.}

We augment this dictionary with word pairs extracted from our small parallel corpora, using standard statistical approaches for lexicon induction. 
Specifically, we employ Fast Align \cite{dyer-etal-2013-simple} on the Bible verses.% to complement the previous dictionary with new entries. 
The expansion of the dictionary with statistical methods follows previous work \cite{dali, zhang2024teaching}. 

Further information about the dictionaries is available in Appendix \ref{sec:app-dict}.

\paragraph{Languages} Because it is difficult to source domain-specific evaluation data in truly low-resource languages, we simulate a low-resource setting, selecting languages not seen during mBART's pretraining.
%Note that while most of these languages are not truly low-resource, the pre-trained model has not seen them during pre-training and they can therefore serve as acceptable substitutes for low-resource languages. One of the main hurdles of finding a more appropriate low-resource language is the availability of data in different domains. 
For the government domain, we experiment on \textbf{Croatian}, \textbf{Icelandic}, \textbf{Maltese}, \textbf{Polish}, and \textbf{Ukrainian}. For the medical domain, we 
%do the same set of experiments for 
use \textbf{Croatian}, \textbf{Icelandic}, \textbf{Maltese}, and \textbf{Polish}. In all cases,  \textbf{English} is our high-resource language.

\section{Experimental Setup} 
\label{sec:exp}
% \apcomment{Add an overview of the experimental approach, explaining the overall shape of the experiments}

%\subsection{Models}
Our goal is to translate from English into our low-resource languages, one at a time. 
In this section, we describe the different approaches we investigate. All of them use mBART as the backbone model 
%for all experiments 
and are implemented 
%all methods 
using
\texttt{fairseq}.\footnote{\url{https://github.com/facebookresearch/fairseq}}

\paragraph{mBART Baseline} Our baseline is
%We use the 
the pretrained mBART model, which has been trained on 25 languages and
%purportedly 
is said to generalize well to unseen languages \cite{liu-etal-2020-multilingual-denoising}.

\paragraph{DALI} We adapt the method from \citet{dali}, who extract a lexicon by mapping word embeddings from the source to the target language. They then use this lexicon to back-translate from the target monolingual data, by word-for-word replacement. The resulting texts are the pseudo-parallel data that are used for training. 
%do not perform the lexicon induction as we 
We produce pseudo-parallel data using the same method, but use the dictionary described in Section \ref{sec:data}.
As we have access to monolingual texts in the \textit{source} language, we do forward-translation instead of back-translation.
%\apedit{instead} assume the lexicon is given, \apedit{make word-for-word replacements,} and we do forward-translation. \apmargincomment{Please correct as necessary - what was missing here is a description of what part of this model we actually use.}
%as we only have access to source-side monolingual data. 

\paragraph{LeCA} \citet{leca} append suggestions to the input to be used in the output. Their model uses a pointer-generator module to potentially copy from the input. Since the model updates just the probability of the next token by also considering copying from the input tokens, it is not a hard constraint. 
%As we want to use our dictionary, 
We match their \textsc{Dictionary Constraint} setting, where suggestions are made by looking up source-side terms in a given dictionary. We implement this on top of the base mBART model.
%as opposed to the basic Transformer. 
Note that LeCA was not originally proposed for low-resource scenarios, and they do not use a pretrained model, instead training the base Transformer model from scratch.

\paragraph{CPT} \citet{liu-etal-2021-continual} continue pretraining mBART on mixed-language text, modifying the pretraining scheme of the model. They corrupt the text by replacing some terms with their translation in the new language, and the model is trained to reconstruct the original text. In our setting, we must use source-side monolingual text only, matching their \textsc{CPT w/ MLT (Src)} method. Note that in our experiments we translate from the high-resource language to the low-resource.%, and if it were the opposite and we were translating to the high-resource language, then using the target-side high-resource monolingual text for pre-training would correspond to their \textsc{CPT w/ MLT (Tgt)}, their main proposed method. 
%\apmargincomment{this last sentence can come out if we need space.}

\paragraph{Combined} We experiment with merging the above methods: first, we  pretrain the model (CPT) and then train it with pseudo-parallel data (DALI) while using pointer-generators (LeCA). 

%\subsection{Metrics}
\paragraph{Metrics }
We evaluate all methods on the test data decribed in Section \ref{sec:data}, using BLEU \cite{papineni-etal-2002-bleu} and ChrF \cite{popovic-2015-chrf} as implemented by sacreBLEU \cite{post-2018-call}. We consider ChrF our main metric, as it focuses on characters and is more informative when translating into morphologically rich languages.%, such as our low-resource languages.
 
% We train mBART on the available resources -  parallel data (the Bible) and monolingual data (in the source language) - in different ways before finally testing it on a test dataset. All the following methods make use of a bilingual lexicon in some capacity. We then analyze the efficacy of different approaches for the task. 

% \noindent \textbf{Baselines}

\begin{table*}[t]
    \centering
    \footnotesize
    \begin{tabular}{cc|c|c|c|c|c|c|c|c|c|c|c} 
    & & \multicolumn{2}{c|}{Croatian} & \multicolumn{2}{c|}{Icelandic} & \multicolumn{2}{c|}{Maltese} & 
    \multicolumn{2}{c|}{Polish} &
    Ukrainian & \multicolumn{2}{c}{Average}
    \\
     & Metric & Gov. & Med. & Gov. & Med. & Gov. & Med. & Gov. & Med. & Gov. & Gov. & Med. \\
     \midrule
    \multirow{2}{*}{mBART} & BLEU & 0.69 & 1.7 & 0.76 & 1.46 & 1.57 & 1.68 & 0.34 & 0.33 & 0.9 & 0.85 & 1.29 \\ 
    & ChrF & 17.34 & 18.62 & 18.97 & 17.72 & 21.61 & 19.42 & 19.11 & 17.37 & 17.83 & 18.97 & 18.28 \\\midrule

    \multirow{2}{*}{DALI} & BLEU & \underline{4.1} & \underline{12.74} & \underline{5.76} & \underline{13.89} & \underline{7.92} & 16.68 & \underline{4.21} & 10.57 & \underline{6.8} & \underline{5.76} & \underline{13.47} \\ 
    & ChrF & 38.87 & \textbf{43.32} & 36.02 & \textbf{41.07} & \textbf{49.55} & 48.77 & \textbf{36.33} & \textbf{36.73} & \textbf{37.51} & \textbf{39.66} & \textbf{42.47} \\\midrule

    \multirow{2}{*}{LeCA} & BLEU & 0.65 & 1.68 & 0.98 & 0.24 & 1.41 & 1.5 & 0.35 & 0.41 & 0.79 & 0.84 & 0.96 \\ 
    & ChrF & 17.48 & 18.23 & 19.24 & 15.97 & 20.6 & 18.56 & 17.6 & 17.11 & 18.74 & 18.73 & 17.47 \\\midrule

    \multirow{2}{*}{CPT} & BLEU & 2.62 & 8.02 & 3.66 & 5.26 & 2.18 & 5.38 & 1.57 & 5.73 & 4.38 & 2.88 & 6.1 \\ 
    & ChrF & 20.46 & 25.19 & 20.67 & 20.56 & 20.42 & 21.86 & 19.19 & 21.03 & 12.35 & 18.62 & 22.16 \\\midrule

    \multirow{2}{*}{Combined} & BLEU & 3.87 & 12.21 & 5.63 & 13.4 & 7.14 & \underline{16.75} & 3.82 & \underline{10.67} & 6.69 & 5.43 & 13.26 \\ 
    & ChrF & \textbf{39.93} & 42.11 & \textbf{36.33} & 40.56 & 48.17 & \textbf{48.88} & 35.72 & 36.11 & 36.46 & 39.32 & 41.92 \\\bottomrule

    \end{tabular}
    \caption{Performance on the all the test sets for the target domains government (Gov.) and medical (Med.) documents. Best BLEU score per column is underlined, while the best ChrF score is indicated in bold.  }
    \label{tab:final-res}
\end{table*}

\section{Results and Discussion}
The results for all languages and domains
%are illustrated 
appear in Table \ref{tab:final-res}. On average, DALI performs best 
%outperforms the other methods 
in the majority of the experiments. It is also the simplest of the methods to implement, as it is model-agnostic and only the training data is manipulated.

LeCA does not help in most cases, supporting
%which is in agreement with
\citet{bafna-etal-2024-pointer}, who observe that pointer-generators are not consistently helpful for low-resource NMT. 
%As mentioned, 
LeCA was not initially devised for low-resource settings, and also the dictionary here includes just one translation per term, with no guarantee of matching the intended target side meaning. %for a single term which may not be the one used on the target side. 
%\citet{maheshwari2022dictdis} address this issue by adding a ``disambiguation network'' that helps the model choose the right candidate among the suggestions.
Since mBART was not pretrained on these languages, its embeddings of words in the low-resource language might not as directly correspond to their source-side, high-resource counterparts; we hypothesize this may be another reason LeCA performs poorly for resource-constrained scenarios. 

CPT is helpful in most of the experiments when compared to plain mBART, but not compared to DALI. After pretraining, the model is fine-tuned only on Bible data. In the pretraining, we reconstruct the source side, so the model only learns to output in the target language from the Bible verses. 
%Even though the 
Pretraining helps the model get more familiar with the domain and establish connections between the embeddings of target words and their respective translations in the high-resource language.%, it still fell short of DALI in terms of boost in performance. 
\footnote{According to \citet{liu-etal-2021-continual}, the performance boost is expected to increase if we have monolingual texts in the target language instead and can use them during pretraining.}% - their \textsc{CPT w/ MLT (Tgt)} method. Even though it was their main proposed method, it does not apply to our setting.}

Combining the methods together 
shows some improvements over other individual methods, but generally
%does not always boost the performance over the individual methods; in fact in most cases it 
fails to reach DALI's performance. Note that the same dataset was used to both pretrain the model (the CPT part) and to then make pseudo-parallel data (the DALI part). 
%\amcomment{Although we decided to use all the data in every step possible, it's worth noting that in our early experiments using different datasets for pretraining and DALI lead to additional gains on top of just using DALI.} 
Since LeCA is not helpful when added to the basic mBART, we also 
%wanted to test the rigor of the CPT+DALI model; which is 
test performance of \textit{Combined}  without LeCA, on the medical domain. 
%The results are shown in Table \ref{tab:cpt-dali}. 
The results (Table~\ref{tab:cpt-dali}) indicate that -- when using the same dataset for both -- adding pretraining on top of DALI can be detrimental, but removing LeCA increases performance
%mostly does not increase performance and can be detrimental in some cases. CPT + DALI also outperforms Combined 
on all languages for the medical domain.

LeCA only uses the dictionary in the final stage, and CPT uses the monolingual data during pretraining, before being fine-tuned on the bible data. DALI and \textit{Combined} are the only methods that have access to source-side target-domain monolingual data during the final stage of training, which could partially explain their superior performance. That a simple method like DALI -- that mostly keeps the word order of the sentence language -- should be the best performing method hints at the extensive room for growth in future work. 

% \amcomment{
% While government data did result in lower BLEU scores, the ChrF scores are usually pretty comparable. ChrF is a better measure than BLEU for estimating output quality for data with richer morphology, because it is more robust to outputs that have the right stem but the wrong morphological inflection. This opens the question of whether there might be differences in morphological complexity across domains, which is something to be investigated in future work.
% }

\begin{figure*}[h!]
    \centering
    \begin{minipage}{0.515\textwidth} % Adjust the width of the image container
        \centering
        \includegraphics[width=\textwidth]{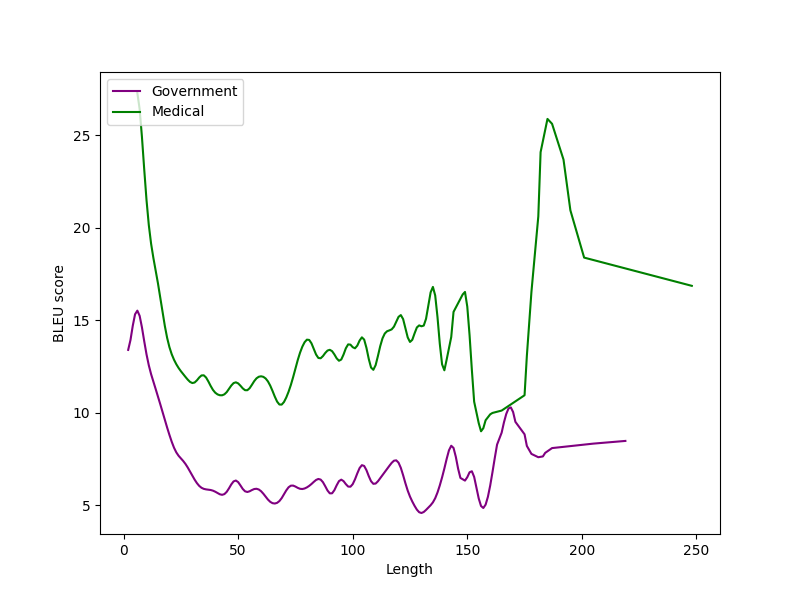}
    \end{minipage}
    \hspace{-0.05\textwidth} % Add some space between the images
    \begin{minipage}{0.515\textwidth}
        \centering
        \includegraphics[width=\textwidth]{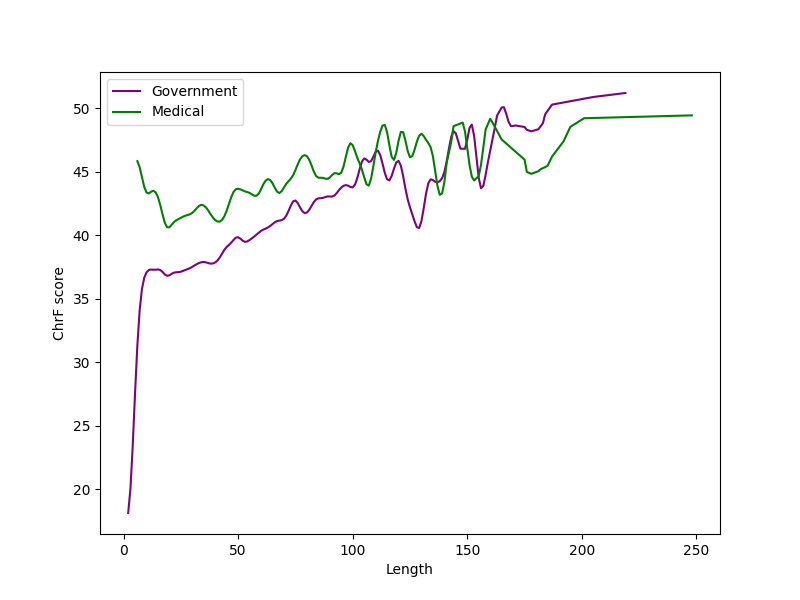}
    \end{minipage}
    \caption{The trend of averaged sentence-level BLEU (left) and ChrF (right) scores against the token length of the reference translation for DALI models. The scores are averaged across all the model outputs  of the same length -- including averaging across languages, where relevant.}
    \label{DALI_scores}
\end{figure*}

Figure \ref{DALI_scores} shows averaged sentence-level BLEU and ChrF scores plotted against their respective reference token lengths for DALI models. For length $l$, we average the scores of the models for different languages if the reference translation is of length $l$. 
% The scores of the models for different languages are averaged for reference translations of the same length . 
We can see that generally the scores seem to get higher with longer sentences, especially for ChrF. 
% The gap between domains is less severe for longer sentences in ChrF, but is pretty consistent for BLEU.

% \begin{figure*}[h!]
%     \centering
%     \includegraphics[width=0.6\textwidth]{tables/DALI_BLEU.png} % Adjust width as needed
%     \caption{This is a caption for the figure.}
%     \label{fig:example} % Optional label for referencing
% \end{figure*}

\paragraph{Example } We see some interesting trends in the outputs. Table~\ref{tab:example-mlt} showcases an example with the outputs of different methods for one sentence from 
the Maltese-medical test set, the language--domain pair with the most significant performance boost. \emph{Warning: these outputs could include distressing language against women
that may harm some readers.} 
Both mBART and LeCA translate in a religious tone. The same is true for CPT, which also tends to copy words from the input -- as it was a part of the reconstruction procedure during pretraining.
It is important to emphasize that Maltese is a morphologically rich language, and the inflections are mostly discarded in the outputs of DALI and \textit{Combined};
for example the words are more likely to be disjoint in their outputs than they are in the target (the first ``\textit{il o$\hbar$ra}'' vs ``\textit{l-o$\hbar$ra}'' in the target), or they can be in different forms (``\textit{huwa}'' vs ``\textit{hija}'').
Note that \textit{ointment} was translated to \textit{infusion} by DALI. Given the sensitivity of the domain, a translation like this can potentially be harmful.

\setlength{\abovecaptionskip}{4pt}
\setlength{\belowcaptionskip}{4pt} % Space below the caption
\begin{table*}[!ht]
\centering
\begin{tabular}{p{1.6cm}|p{5.5cm}|p{1.6cm}|p{5.5cm}}
\multicolumn{4}{c}{Source: if the other eye medicine is an eye ointment it should be used last } 
\\\hline
 mBART: & jekk il-mara l-ie$\hbar$or hi çajn o$\hbar$ra , hi çandha tinçatalha l-a$\hbar$$\hbar$ar fl-a$\hbar$$\hbar$ar  & LeCA: & inkella jekk il-mara l-ie$\hbar$or hi żejt , tkun maçmula l-a$\hbar$$\hbar$ar  \\ \cline{2-2} \cline{4-4}
BT:& if the other woman is someone else, she should be punished in the end  & BT:& otherwise if the other woman is a virgin, she will be the worst \\
 \hline
DALI: & jekk il o$\hbar$ra g$\hbar$ajn mediċina huwa an g$\hbar$ajn infużjoni dan g$\hbar$andu tkun użati l-a$\hbar$$\hbar$ar  & CPT: & jekk l-o$\hbar$rajn ta ' l-ie$\hbar$or hi çajnejja ointment , it should be used l-a$\hbar$$\hbar$ar\\  \cline{2-2} \cline{4-4}
BT: & If the other eye medicine is an eye infusion, this should be used last & BT: & Even though the other one is a çajnejja ointment, it should be used last \\
\hline
Combined: & jekk il o$\hbar$ra g$\hbar$ajn mediċina huwa an g$\hbar$ajn ointment dan g$\hbar$andu tkun użati l-a$\hbar$$\hbar$ar  & Target: & jekk il-mediċina tal-g$\hbar$ajnejn l-o$\hbar$ra hija ingwent tal-g$\hbar$ajnejn , dan g$\hbar$andu jintuża l-a$\hbar$$\hbar$ar  \\ \cline{2-2} \cline{4-4} 
BT: & If the other eye medicine is an eye ointment, this should be used last & BT: & Although the other eye medicine is an eye ointment, this should be used last \\
\hline
% \multicolumn{3}{}{} Output & Prevenar is a medicine containing the design of Arixtra .              \\ \hline
\end{tabular} 
\caption{\textbf{Warning: this table contains harmful language about women that may distress some readers.} An example of different model outputs for a Maltese sentence in the medical domain. For better comparison, the back-translations (BT) of the outputs to English are also included, done via Google Translate.}
\label{tab:example-mlt}
\end{table*}

Maltese at times has a different word order than that of English (
``\textit{il-mediċina  tal-g$\hbar$ajnejn}'' is translated as ``\textit{g$\hbar$ajn mediċina}'', which matches the order  of its English counterpart ``eye medicine''), and it is also more flexible. DALI and \textit{Combined} produce word orders 
that closely follow the source language.

\paragraph{Human Evaluation} Conducting a small-scale human evaluation of the Polish government translations of 25 source sentences, we find that, while DALI improves the communication of the overall semantics of the sentences, there is certainly room for improvement, especially when it comes to fluency and generating grammatical output.
Additional model outputs and details of the human evaluation can be found in Appendix \ref{app:example}.

%\ammargincomment{I feel like I'm saying word order a lot. It'd be nice if someone could paraphrase this!}

%We thereby aim to call for more attention to this specific setting that covers the case of many low-resource languages.

\section{Conclusion}

This paper introduces a realistic setting that has been previously overlooked: DA for NMT into a low-resource from a high-resource language, with available resources restricted to limited parallel text, a dictionary, and monolingual texts in the high-resource language. 
%We tested some existing methods that could be applied to this task, all of which involve the dictionary. 
%\st{Our results indicate that for the languages under study, these methods mostly fail to boost the performance to an acceptable level that would be considered useful.} \amcomment{
%We show that
%, for the languages under study, 
The simplest approach -- DALI -- yields the best results, more than doubling baseline performance. A small-scale human evaluation indicates ample room for improvement, and
we advocate for increased focus on this setting.% within the NLP community.
%, particularly for languages with data restricted to specific domains, as these languages are the most in need of domain adaptation and are often overlooked. 

\section*{Limitations}

It is important that these experiments be conducted for truly low-resource languages. The scope of this work was limited due to the availability of datasets in different domains for such resource-constrained languages; which was the main reason we resorted to experimenting on simulated low-resource languages. 
Limitations of finding domain-specific corpora for low-resource languages also extend to finding domain-specific dictionaries, and our dictionaries prepared with Google Translate only mimic target-domain dictionaries. 
In addition, we based our experiments on mBART only, and we leave the study of other multilingual pretrained models and LLMs (or even smaller, non-pretrained models like the base Transformer) in this setting for future work. 
% \apcomment{You should also talk about limitations based on the dictionary procedure.}

\section*{Ethics Statement}

As our research shows that these methods do not sufficiently enhance performance for the models to be deemed useful, there are some caveats to be mindful of. Specifically, these methods should not be used for real-world MT in critical contexts involving low-resource languages; e.g. providing medical advice based on the translations produced by the model. \\
All the data used in the study is 
publicly available
%available to the public 
(see Appendix \ref{sec:app-data}).

\section*{Acknowledgments}

We sincerely thank the anonymous reviewers for their helpful feedback. We also wish to express our gratitude to LECS and NALA labs at the University of Colorado Boulder. This work made use of the Blanca condo computing resource at the University of Colorado Boulder, which is supported through joint funding by computing users and the university.

\bibliography{main}

\appendix

\newpage
\section{Data}

\subsection{Datasets}
\label{sec:app-data}
Here are the details for the data used in training, testing and potential pretraining and pseudo data generation.  All the datasets are lower-cased.

\subsubsection{Parallel Data}

The parallel data come from the New Testament verses from the Johns Hopkins University Bible Corpus \cite{mccarthy-etal-2020-johns}. For all experiments, 8\% of the verses are extracted to be used as validation data.
%validation data is extracted from these verses, making up for 8\% of the verses. 
The number of verses per language is in the range 7k-8k. The test dataset is of another domain, and it is discussed in \ref{sec-pseudo}. The sizes of the train and validation datasets for different languages are shown in Table \ref{tab:bible-stat}.\\

\begin{table}[!h]
    \centering
    \begin{tabular}{|l|c|c|} \hline 
      \textbf{Language} & \textbf{Train} & \textbf{Validation} \\\hline
     Croatian & 7290 & 634 \\\hline
     Icelandic & 7167 & 624  \\\hline
     Maltese & 7122 & 620  \\\hline
     Polish & 7293 & 635  \\\hline
     % Tamil & 137 & 10428 & - \\\hline
     Ukrainian & 6799 & 592 \\\hline
     
    \end{tabular}
    \caption{Number of parallel Bible verses used in training and validation across different languages.}
    \label{tab:bible-stat}
\end{table}

\subsubsection{Pseudo-Parallel Data}
\label{sec-pseudo}

We use the Tilde MODEL corpus \cite{rozis-skadins-2017-tilde} for the majority of our experiments, as it is listed as an available resource for many of WMT tasks during the last few years \cite{barrault-etal-2020-findings, akhbardeh-etal-2021-findings, kocmi-etal-2022-findings, kocmi-etal-2023-findings}. In all experiments, we retain 1500 sentence pairs for testing. This is the only portion for which we keep the target side, as we only manipulate the source side from the rest of them. 
In case there are more than 200K available pairs, we use seed $= 42$ to randomly choose 200K pairs from the dataset. \\

\noindent \textbf{Government Domain} \\

\textbf{Croatian}: We use EESC from the Tilde MODEL, that comprises document texts from the ``European Economic and Social Committee'' document portal. The full 200K sentence pairs are used for training and pretraining. 

\textbf{Icelandic}: We use the concatenation  of the following three datasets: ``Government Offices in Iceland - Reports'', ``Government Offices in Iceland – Legislation and regulations'', and ``Bilingual English-Icelandic parallel corpus from the official Nordic cooperation website'' from the European Language Resource Coordination.\footnote{\href{https://language-data-space.ec.europa.eu/related-initiatives/elrc_en}{https://language-data-space.ec.europa.eu/related-initiatives/elrc\_en}} This makes for a dataset of size 87233 that is used for both training and pretraining. 

\textbf{Maltese}: As was the case with Croatian, we use the English-Maltese subsection of EESC, and we choose 200K sentence pairs from all available pairs. 

\textbf{Polish}: We utilize RAPID from the Tilde MODEL, composed of the press releases of ``Press Release Database of European Commission'' released between 1975 and the end of 2016. 200K sentence pairs are extracted.

\textbf{Ukrainian}: We use ``EU acts in Ukrainian'' from the European Language Resource Coordination, resulting in 116,568 sentence pairs. \\
\\
\noindent \textbf{Medical Domain}

For the four languages investigated (Croatian, Icelandic, Maltese, Polish), we use EMA from the Tilde MODEL. It is compiled from texts available via the European Medicines Agency document portal. All of these languages had more than 200K sentence pairs, from which 200K were extracted.

\subsection{Dictionaries}
\label{sec:app-dict}
The method with which the dictionaries are composed is described in \ref{dict:making}. Since many of the lemmas might have several inflected forms that appear in the text, the dictionary sizes are larger than 5000, usually varying between 8k-10k. Here are the exact size of the dictionaries. In Table \ref{tab:dict-stat}, the column `Bible' denotes the number of terms extracted from the Bible and added to the \textit{in-domain} terms that were drawn out from the monolingual source-side corpus. Note that if the term already exists in the in-domain dictionary, we do not replace it with the one from the Bible. The columns `Government' and `Medical' indicate the final size of the dictionary of their respective domains, including the new terms from the Bible. 

\begin{table*}[!h]
    \centering
    \begin{tabular}{|l|c|c|c|} \hline
     & \textbf{Bible} & \textbf{Government} & \textbf{Medical} \\\hline
     Croatian & 182 & 9948 & 8142 \\\hline
     Icelandic & 359 & 10004 & 8383 \\\hline
     Maltese & 319 & 10004 & 8309 \\\hline
     Polish & 417 & 10192 & 8337 \\\hline
     % Tamil & 137 & 10428 & - \\\hline
     Ukrainian & 283 & 9437 & - \\\hline
     
    \end{tabular}
    \caption{Sizes of different dictionaries used for different languages and domains.}
    \label{tab:dict-stat}
\end{table*}

\section{Training}

The details of training are as follows. Each setting was trained once, and the experiments were done on NVIDIA A100 GPUs. \\
Note that some of the experiments rely on others; for example, \textit{Combined} has three stages of updating the model: 1) continual pretraining on the domain-specific texts, 2) training the model from step 1 on the Bible dataset (CPT), 3) training the model from step 2 on the pseudo parallel data + Bible (which is the DALI part). 
Some notable libraries we use include: 
\begin{itemize}
    \item fairseq v0.12.2 (which we modified to run our methods)
    \item torch v1.13.1
    \item sentencepiece v0.1.99
    \item transformers v4.30.2. 
\end{itemize}    

\subsection{Training Hyperparameters}

We implemented LeCA and CPT on \texttt{fairseq} for mBART, and had to change parts of the main library for compatibility. Since mBART needs a language id, we added new tokens for these new languages. We initialized their embeddings randomly (following the method for parameter initialization in \citet{liu-etal-2020-multilingual-denoising}).
% As a result, some hyperparameters are added to the usual \texttt{fairseq} ones, and they are not listed here\footnote{They are explained in the code base.}.\\
The \texttt{fairseq} hyperparameters used in pretraining and training are listed in Table \ref{tab:hyp}.

\begin{table*}[h]
    \centering
    \begin{tabular}{|l|c|c|}
        \hline
        \textbf{Hyperparameter} & \textbf{Pretraining} & \textbf{Training}\\
        \hline
        arch & \multicolumn{2}{c|}{mbart\_large} \\\hline
        lr-scheduler & \multicolumn{2}{c|}{polynomial\_decay} \\\hline
        lr & \multicolumn{2}{c|}{3e-5} \\\hline
        optimizer & \multicolumn{2}{c|}{adam} \\\hline
        adam-eps & \multicolumn{2}{c|}{1e-06} \\ \hline
        adam-betas & \multicolumn{2}{c|}{(0.9, 0.98)} \\ \hline
        dropout & \multicolumn{2}{c|}{0.3} \\ \hline
        attention-dropout & \multicolumn{2}{c|}{0.1} \\ \hline
        bpe & \multicolumn{2}{c|}{sentencepiece} \\\hline
        max-tokens & \multicolumn{2}{c|}{1024} \\\hline
        save-interval & \multicolumn{2}{c|}{5} \\\hline
        criterion & \multicolumn{2}{c|}{label\_smoothed\_cross\_entropy} \\\hline
        no-epoch-checkpoints & \multicolumn{2}{c|}{True} \\\hline
        layernorm-embedding & \multicolumn{2}{c|}{True} \\\hline
        encoder-normalize-before & \multicolumn{2}{c|}{True} \\\hline
        decoder-normalize-before & \multicolumn{2}{c|}{True} \\\hline
        share-decoder-input-output-embed & \multicolumn{2}{c|}{True} \\\hline
        encoder-learned-pos & \multicolumn{2}{c|}{True} \\\hline
        required-batch-size-multiple & \multicolumn{2}{c|}{1} \\\hline
        label-smoothing & \multicolumn{2}{c|}{0.2} \\\hline
        update-freq & \multicolumn{2}{c|}{2} \\\hline
        seed & \multicolumn{2}{c|}{42} \\\hline
        warmup-updates & 2000 & 1000 \\\hline
        min-epoch & 20 & 75 \\\hline
        min-epoch & 60 & 150 \\\hline
        patience & 10 & 50 \\\hline
        total-num-update  & \multicolumn{2}{c|}{(number of steps in one epoch) * max-spoch} \\\hline
        task & denoising & translation\_from\_pretrained\_bart \\\hline
        mask & 0.35 & - \\\hline
        tokens-per-sample & 384 & - \\\hline
        poisson-lambda & 3.5 & - \\\hline
        mask-length & span-poisson & - \\\hline
        replace-length & 1 & - \\\hline
        rotate & 0 & - \\\hline
        permute-sentences & 0 & - \\\hline
        
        \hline
    \end{tabular}
        \caption{Pretraining and training hyperparameters}
    \label{tab:hyp}
\end{table*}

\subsection{Batches for experiments with DALI}
In experiments containing DALI - DALI, \textit{Combined} (and CPT + DALI which is done for medical domain experiments, as presented in Table \ref{tab:cpt-dali}) - batches are constructed in a particular way. 

In each batch, we have the same number of instances from out-of-domain parallel data and in-domain pseudo-parallel data. Training batches do not contain overlapping in-domain pseudo-parallel data, but  
%In their experiments, however, these out-of-domain parallel sentences were not necessarily limited like ours, as we only have Bible verses as the parallel data. So here, 
we do use the same out-of-domain parallel data in every batch, because we are limited to Bible verses for parallel data.
%but we still have different in-domain pseudo-parallel data. 

\section{Additional Outputs and Evaluation}
\label{app:example}

% \input{tables/maltExample}

% Table~\ref{tab:example-mlt} is discussed in Section~\ref{sec:exp}.

Table \ref{tab:example-pol} shows the model outputs for an example sentence from the Polish test set in the government domain. We can see the same patterns of religious phrasing in mBART, LeCA and CPT.
Some words have different translations than those used in the target translation; e.g. the model translates \textit{banking} as ``bankowość'' while
``bankowej'' is used in the reference.
Polish is also a morphologically rich language and it sometimes does not match English's word order. Here, for example, \textit{banking union} should be translated as ``unii bankowej'' while in DALI and 
and \textit{Combined} the phrase is translated as ``bankowość unia'', in the same order as in the English sentence. 

\paragraph{Human evaluation} We conduct a small-scale human evaluation on a set of 25 randomly-selected sentences from the test set of Polish government data. A Polish native speaker annotator scored the translations for both communication of the \textbf{intended meaning} and the correctness of the overall \textbf{grammatical structure}, using a scale from 0 to 5.
%(from 0 to 5) to the translations for both the communication of the semantics and the credibility of the overall grammatical structure. 
Only the translations of the original mBART (baseline) and DALI are compared. The average scores for \textbf{meaning} for baseline and DALI were 0.12 and 0.2, respectively. For the \textbf{grammar}, both models were given an average score close to 0. (Perhaps not surprising for a language with the morphological richness of Polish.)
%got 0 which in the case of Polish, is understandable as it is a morphologically rich language. 
Of course, more in-depth study of the results is needed to draw any strong conclusions about usability.

\paragraph{Output statistics} The average number of words, number of tokens, and number of characters of the outputs of different methods against the reference translations are presented in Table \ref{tab:methods-stats}. For number of words, an output is split by white-spaces. For tokens, the mBART tokenizer is used. We average the results across languages. We report the averages because relative length patterns tend to be consistent across languages. The full table containing language specific statistic is available on the GitHub repository: \href{https://github.com/alimrsn79/da_lr_nmt/}{https://github.com/alimrsn79/da\_lr\_nmt}.

\begin{table*}[t]
\centering
\begin{tabular}{p{1.4cm}p{5.5cm}|p{0.9cm}p{5.5cm}}
\multicolumn{4}{c}{Source: in the banking union , those funds are pooled together gradually .} 
\\\hline
 mBART: & przetoż zgromadzi się wszystkie , które są w łodzi . & LeCA: & przetoż zgromadzi one członki w lichwiarze . \\ \hline
DALI: & w the bankowość unia , te fundusze czy poszczepiony razem stopniowo . & CPT: & w banking union wespół to zgromadziło , i nader to zgromadziło .   \\ \hline
Combined: & w the bankowość unia , te fundusze czy pooled razem stopniowo . & Target: & fundusze te będą gromadzone stopniowo w ramach unii bankowej . \\ \hline
% \multicolumn{3}{}{} Output & Prevenar is a medicine containing the design of Arixtra .              \\ \hline
\end{tabular} 
\caption{An example of different model outputs for a Polish sentence in the government domain.}
\label{tab:example-pol}
\end{table*}

\begin{table*}[t]
    \centering
    \footnotesize
    \begin{tabular}{cc|c|c|c} 
    & & \multicolumn{3}{c}{Average}
    \\
     & Domain & Words & Tokens & Characters \\
     \midrule
     \multirow{2}{*}{Reference} & Gov. & 23.51 & 44.65 & 154.55  \\ 
    & Med. & 19.65 & 39.43 & 120.42 \\\midrule
    
    \multirow{2}{*}{mBART} & Gov. & 25.12 & 49.25 & 132.7  \\ 
    & Med. & 20.91 & 44.27 & 110.94 \\\midrule

    \multirow{2}{*}{DALI} & Gov. & 26.83 & 43.99 & 155.28  \\ 
    & Med. & 20.3 & 37.38 & 115.7 \\\midrule

    \multirow{2}{*}{LeCA} & Gov. & 24.82 & 49.15 & 133.92  \\ 
    & Med. & 22.94 & 46.02 & 117.94 \\\midrule

    \multirow{2}{*}{CPT} & Gov. & 26.44 & 47.85 & 139.7  \\ 
    & Med. & 20.5 & 38.23 & 109.46 \\\midrule

    \multirow{2}{*}{Combined} & Gov. & 26.93 & 43.97 & 155.95  \\ 
    & Med. & 20.31 & 37.41 & 116.09 \\\bottomrule
    
    \end{tabular}
    \caption{The average number of words, tokens, and characters of the outputs of different methods against the reference translation. The results are averaged over all the experiments. }
    \label{tab:methods-stats}
\end{table*}

\begin{table*}[!h]
    \centering
    \begin{tabular}{cc|c|c|c|c|c} 
    & Metric & Croatian & Icelandic & Maltese & 
    Polish & Average \\
     \hline
    \multirow{2}{*}{DALI} & BLEU & \underline{12.74} & \underline{13.89} & 16.68 & 10.57 & \underline{13.47} \\ 
    & ChrF & \textbf{43.32} & \textbf{41.07} & 48.77 & \textbf{36.73} & \textbf{42.27} \\\hline

    \multirow{2}{*}{Combined} & BLEU & 12.21 & 13.4 & 16.75 & 10.67 & 13.26 \\ 
    & ChrF & 42.11 & 40.56 & 48.88 & 36.11 & 41.92 \\\hline

    \multirow{2}{*}{CPT + DALI} & BLEU & 12.59 & 13.28 & \underline{17.03} & \underline{10.88} & 13.45 \\ 
    & ChrF & 42.6 & 38.67 & \textbf{49.1} & 36.36 & 41.68 \\\hline
    \end{tabular}
    \caption{Comparing CPT + DALI with DALI and \textit{Combined} on the medical domain.}
    \label{tab:cpt-dali}
\end{table*}

\begin{table*}[!h]
    \centering
    \begin{tabular}{cc|cc} 
    & & \multicolumn{2}{c}{Icelandic} \\
    & Metric & Gov. & Med. \\
     \hline
     
    \multirow{2}{*}{DALI} & BLEU & 5.76 & 13.89\\ 
    & ChrF & 36.02 & 41.07 \\\hline

    \multirow{2}{*}{Combined} & BLEU & 5.63 & 13.4\\ 
    & ChrF & 36.33 & 40.56 \\\hline

    \multirow{2}{*}{Full} & BLEU & 34.46 & 55.98\\ 
    & ChrF & 59.1 & 74.05 \\\hline
    \end{tabular}
    \caption{Comparing the model trained on the full parallel dataset with DALI and \textit{Combined} that only had access to the source side, for Icelandic. The full models were trained with the same hyperparameters as the training column in Table \ref{tab:hyp}, but the training was done on the full in-domain parallel text instead of the Bible and pseudo-parallel sentences.}
    \label{tab:ice-full}
\end{table*}

\end{document}